\documentclass[11pt,a4paper]{article}
\usepackage[hyperref]{acl2018}
\usepackage{times}
\usepackage{latexsym}

\usepackage{url}
\usepackage{latexsym}
\usepackage{tabularx}
\usepackage{xcolor}
\usepackage{graphicx}

\aclfinalcopy 

\usepackage[T2A,T1]{fontenc}
\usepackage[utf8]{inputenc}
\usepackage[russian,english]{babel}

\usepackage{CJK,algorithm,algorithmic,amssymb,amsmath,array,epsfig,graphics,multirow,array,float,subfigure,verbatim,epstopdf}
\usepackage{enumitem}
\usepackage{color}
\usepackage{tabularx}
\usepackage{hhline}
\usepackage{multirow}

\title{Describing a Knowledge Base}

\usepackage{xparse}
\NewDocumentCommand{\zhiying}{ mO{} }{\textcolor{orange}{\textsuperscript{\textit{Zhiying}}\textsf{\textbf{\small[#1]}}}}
\NewDocumentCommand{\qingyun}{ mO{} }{\textcolor{purple}{\textsuperscript{\textit{Qingyun}}\textsf{\textbf{\small[#1]}}}}
\NewDocumentCommand{\heng}{ mO{} }{\textcolor{red}{\textsuperscript{\textit{Heng}}\textsf{\textbf{\small[#1]}}}}
\NewDocumentCommand{\lifu}{ mO{} }{\textcolor{pink}{\textsuperscript{\textit{Lifu}}\textsf{\textbf{\small[#1]}}}}

\definecolor{fl1}{RGB}{246,119,118}
\definecolor{fl2}{RGB}{250,184,146}
\definecolor{fl3}{RGB}{255,249,174}
\definecolor{fl4}{RGB}{184,217,200}
\definecolor{fl5}{RGB}{114,185,226}
\definecolor{fl6}{RGB}{255,179,186}
\definecolor{fl7}{RGB}{255,223,186}
\definecolor{fl8}{RGB}{255,255,106}
\definecolor{fl9}{RGB}{186,255,201}
\definecolor{fl10}{RGB}{186,225,25}

\author{
Qingyun Wang$^{1}$, \ Xiaoman Pan$^{1}$, \ Lifu Huang$^1$, \ Boliang Zhang$^1$,\ \textbf{Zhiying Jiang}$^1$, \\ \ \textbf{Heng Ji}$^1$, \ \textbf{Kevin Knight}$^2$ \\
$^{1}$ Rensselaer Polytechnic Institute \\
{\tt jih@rpi.edu} \\
$^{2}$ DiDi Labs and University of Southern California \\
{\tt knight@isi.edu}
}
\date{}

\begin{document}

\maketitle

\begin{abstract}




We aim to automatically generate natural language descriptions about an input structured knowledge base (KB). We build our generation framework based on a pointer network which can copy facts from the input KB, and add two attention mechanisms: (i) \emph{slot-aware attention} to capture the association between a slot type and its corresponding slot value; and (ii) a new \emph{table position self-attention} to capture the inter-dependencies among related slots. For evaluation, besides standard metrics including BLEU, METEOR, and ROUGE, we propose a \textit{KB reconstruction} based metric by extracting a KB from the generation output and comparing it with the input KB. We also create a new data set which includes 106,216 pairs of structured KBs and their corresponding natural language descriptions for two distinct entity types.
Experiments show that our approach significantly outperforms state-of-the-art methods. The reconstructed KB achieves 68.8\% - 72.6\% F-score.\footnote{We make all data sets and programs of various models publicly available for research purposes at \url{https://github.com/EagleW/Describing_a_Knowledge_Base}.} 



\end{abstract}



\section{Introduction}
\label{secintro}

\emph{Show and tell}, showing an audience something and telling them about it, is a common classroom activity for early elementary school kids. As a similar practice for 
knowledge propagation, we often need to describe and/or explain the information in a structured knowledge base (KB) 
in natural language, in order to make the knowledge elements and their connections easier to comprehend. For example, \cite{cawsey1997natural} presents a natural language generation system to convert structured medical records to natural language text descriptions, which enables more effective communication between health care providers and their patients and among health care providers themselves.

Moreover, 51\% of entity attributes in the current English Wikipedia Infoboxes are not described in English articles in the Wikipedia dump of April 1, 2018. The availability of vast amounts of Linked Open Data (LOD) and Wikipedia derived resources such as DBPedia, WikiData and YAGO encourages pursuing a new direction of knowledge-driven~\cite{Whitehead18,Lu18imagecaption} or semantically oriented~\cite{Bouayad2013} Natural Language Generation (NLG). 
We aim to fill in this knowledge gap by developing a 
system that can take a KB (consisted of a set of slot types and their values) about an entity as input 
(see example in Table~\ref{tab:example1input}), and automatically generate a natural language description (Table~\ref{tab:example1output}).



\begin{table}[!htb]
\centering
\small
\setlength\tabcolsep{3pt}
\setlength\extrarowheight{3pt}
\begin{tabularx}{\linewidth}{|>{\hsize=1.4\hsize}X|>{\centering\arraybackslash\hsize=.5\hsize}X|>{\hsize=2.0\hsize}X|>{\hsize=0.8\hsize}X|>{\centering\arraybackslash\hsize=.3\hsize}X|}
\hline
\textbf{Slot Type} & \textbf{Row}       & \multicolumn{3}{c|}{\textbf{Slot Value}}          \\ \hline
Name               & 1                  & \multicolumn{3}{c|}{\colorbox{fl1!40}{Silvi Jan}}                 \\ \hline
     & 2 & \multicolumn{3}{c|}{\colorbox{fl5!20}{ASA Tel Aviv University}}  \\ \cline{2-5} 
Member of          & 3 &\multicolumn{3}{c|}{\colorbox{fl5!30}{Hapoel Tel Aviv F.C.(women)}}     \\ \cline{2-5} 
Sports team           & 4 &\multicolumn{3}{c|}{\colorbox{fl5!40}{Maccabi Holon F.C. (women)}}     \\ \cline{2-5} 
                   & \multirow{2}{*}{5} & \colorbox{fl5!50}{Israel women's na-} & Matches & \colorbox{fl6!30}{22}          \\ \cline{4-5} 
                   &                    & \colorbox{fl5!50}{tional football team}         & Goals   & \colorbox{fl7!40}{29}             \\ \hline
Date of Birth           & 6                  & \multicolumn{3}{c|}{\colorbox{fl2!40}{27 October 1973}}             \\ \hline
Country of Citizenship        & 7                  & \multicolumn{3}{c|}{\colorbox{fl4!40}{Israel}}                      \\ \hline
Position        & 8                & \multicolumn{3}{c|}{\colorbox{fl8!40}{Forward (association football)}}                      \\ \hline
\end{tabularx}
\vspace{-2mm}
\caption{Input: Structured Knowledge Base\label{tab:example1input} 
}
\vspace{-2mm}
\end{table}

\begin{table*}[!htb]
\small
\setlength\tabcolsep{4pt}
\setlength\extrarowheight{3pt}
\centering
\begin{tabularx}{\linewidth}{|>{\hsize=.3\hsize}X|>{\hsize=1.7\hsize}X|}
\hline
Reference & \colorbox{fl1!40}{Silvi Jan} (born \colorbox{fl2!40}{27 October 1973}) is a retired female \colorbox{fl4!40}{Israeli}. \colorbox{fl1!40}{Silvi Jan} has been a \colorbox{fl8!40}{Forward (association football)} for the \colorbox{fl5!50}{Israel women's national football team} for many years appearing in \colorbox{fl6!30}{22} matches and scoring \colorbox{fl7!40}{29} goals. After \colorbox{fl5!30}{Hapoel Tel Aviv F.C.(women)} folded, Jan signed with \colorbox{fl5!40}{Maccabi Holon F.C. (women)} where she played until her retirement in 2007. In January 2009, Jan returned to league action and joined \colorbox{fl5!20}{ASA Tel Aviv University}. In 1999, with the establishment of the Israeli Women's League, Jan returned to \colorbox{fl4!40}{Israel} and signed with \colorbox{fl5!30}{Hapoel Tel Aviv F.C.(women)} .\\ \hline
Seq2seq & (born 23 April 1981) is a retired \colorbox{fl4!40}{Israeli} footballer. He played for the Thailand 's (scoring one goal) and was a member of the team that won the first ever player in the history of the National Basketball League. She played for the team from 1997 to 2001 scoring \colorbox{fl7!40}{29} goals. She played for the team from 1997 to 2001 scoring \colorbox{fl7!40}{29} goals. She played for the team from 1999 to 2001 and played for the team in the 1997 and 2003 seasons.\\\hline
Pointer & \colorbox{fl1!40}{Silvi Jan} the fourth past the Maccabi Holon F.C. (women). On 27 October 1973 in 29 2014) (born 22) is a former \colorbox{fl4!40}{Israel}. She was a \colorbox{fl8!40}{Forward (association football)} and currently plays for \colorbox{fl5!30}{Hapoel Tel Aviv F.C.(women)} in the Swedish league. She played for the \colorbox{fl5!20}{ASA Tel Aviv University} in the Swedish league. She was a member of the \colorbox{fl5!50}{Israel women's national football team} at the beginning of the 2008 season. \\\hline
+ Type  & \colorbox{fl1!40}{Silvi Jan} (born \colorbox{fl2!40}{27 October 1973}) is a former \colorbox{fl4!40}{Israeli} footballer. He played for \colorbox{fl5!30}{Hapoel Tel Aviv} \colorbox{fl5!30}{F.C.(women)} and \colorbox{fl5!20}{ASA Tel Aviv University}. \\ \hline
+ Type \& Position & \colorbox{fl1!40}{Silvi Jan} (born \colorbox{fl2!40}{27 October 1973}) is a former \colorbox{fl4!40}{Israel}. He played for \colorbox{fl5!50}{Israel women's nation-} \colorbox{fl5!50}{al football team}, \colorbox{fl5!30}{Hapoel Tel Aviv F.C.(women)}, \colorbox{fl5!20}{ASA Tel Aviv University} and \colorbox{fl5!40}{Maccabi Holon} \colorbox{fl5!40}{F.C. (women)}. He was capped \colorbox{fl6!30}{22} times for the \colorbox{fl5!50}{Israel women's national football team}.\\ \hline
\end{tabularx}
\vspace{-2mm}
\caption{Human and System Generated Descriptions about the KB in Table~\ref{tab:example1input}
\label{tab:example1output}}
\vspace{-2mm}
\end{table*}

\textbf{Neural generation to generalize linguistic expressions.} One major challenge lies in generalizing a wide variety of expressions, patterns, templates and styles which human use to describe the same slot type. For example, to describe a football player's membership with a team, we can use various phrases including \emph{member of}, \emph{traded to}, \emph{drafted by}, \emph{played for}, \emph{face of}, \emph{loaned to} and \emph{signed for}. 
Instead of manually crafting 
patterns for each slot type, we leverage the existing pairs of structured slots from Wikipedia infoboxes and Wikidata~\cite{wiki} 
and the corresponding sentences describing these slots in Wikipedia articles as our training data, to learn a deep neural network based generator.


\textbf{Pointer network to copy over facts.} The previous work~\cite{table2text17} considers the slot type and slot value as two sequences and applies a sequence to sequence (seq2seq) framework~\cite{cho2014learning} for generation. 
However, the task of describing structured knowledge is fundamentally different from creative writing, because we need to cover 
the knowledge elements contained in the input KB, and the goal of generation is mainly to clearly describe the semantic connections among these knowledge elements in an accurate and coherent way. 
The seq2seq model fails to capture such connections and tends to generate wrong information (e.g., \emph{Thailand} in Table~\ref{tab:example1output}).
To address this challenge, we choose a pointer network~\citep{hybridp17} to copy slot values directly from the input KB.

\textbf{Slot type attention.} However, the copying mechanism in the pointer network is not able to capture the alignment between a 
slot type and its slot value, and thus it 
often assigns facts to wrong slots. For example, \emph{22} in Table~\ref{tab:example1output} should be the number of matches instead of birth date. 
It also tends to repeat the same slot value based on language model, e.g.,  
``\textit{Uroplatus ebenaui is a of gecko endemic to \textbf{Madagascar}. The Uroplatus is a member of the species of the genus \textbf{Madagascar}.}''. We propose a \textbf{Slot-aware Attention} mechanism to compute slot type attention and slot value attention simultaneously and capture their correlation. Attention mechanism in deep neural networks~\cite{Denil2012} is inspired from human visual attention, which refers to human's capability to focus on a certain region of an image with high resolution while perceiving the surrounding image in low resolution. It allows the neural network to have access to the hidden state of the encoder, and thus learn what to attend to. For example, for a \emph{Date of Birth} slot type, words such as \emph{born} may receive higher attention than \emph{female}. 
As we can see in Table~\ref{tab:example1output} (\textit{+Type}), the output with slot type attention contains more precise slots. 

\textbf{Table position attention.} 
Multiple slots are often interdependent. For example, a football player may join multiple teams, with each team associated with a certain number of points, goals, scores and games participated. 
We design a new table position based self-attention to capture correlations among interdependent slots and put them in the same sentence. For example, our model successfully associates the number of matches \textit{22} with the \textit{Israel women's national football team} as shown in Table~\ref{tab:example1output}.


The major contributions of this paper are: 
\begin{itemize}
\item For the first time, we propose a new table position attention which proves to be effective at capturing inter-dependencies among facts. This new approach achieves 2.5\%-7.8\% F-score gain at KB reconstruction.
\item We propose a \emph{KB reconstruction based metric} to evaluate how many facts are correctly expressed in the generation output. 
\item We create a large dataset of KBs paired with natural language descriptions for 106,216 entities, which can serve as a new benchmark.  
\end{itemize}

\section{Model}


\begin{figure*}[!htb]
\centering\small
\includegraphics[width=.9\textwidth]{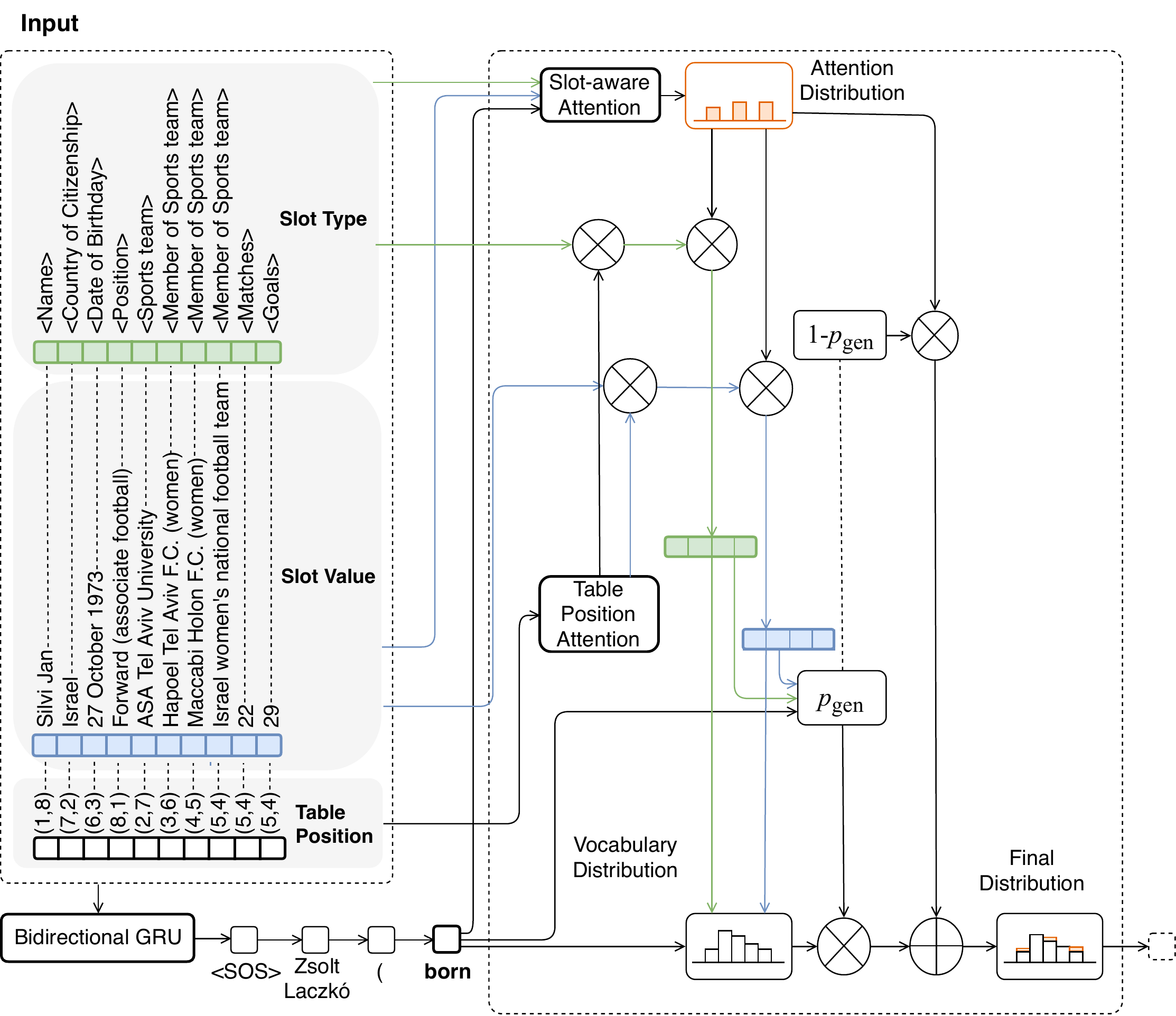}
\caption{KB-to-Language Generation Model Overview}
\label{fig:overview}
\vspace{-2mm}
\end{figure*}

We formulate the input structured KB to the model as a list of triples: 
$L=[(s_1, v_1, (r_1, \hat{r_1})), ..., (s_n, v_n, (r_n, \hat{r_n}))]$, where 
$s_i$ denotes a slot type (e.g., \emph{Country of Citizenship}), $v_i$ denotes the corresponding slot value (e.g., \emph{Israel}), and $(r_i, \hat{r_i})$ denotes the position of the triple in the input list
and consists of the forward position $r_{i}$ and the backward position $\hat{r_{i}}=n-r_{i}+1$
. The outcome of the model is a 
paragraph $Y=[y_1, y_2, ..., y_m]$.
The training instances for the generator are provided in the form of: $T=[(L_1, Y_1), 
..., (L_k, Y_k)]$. 



\subsection{Sequence-to-Sequence with Slot-aware Attention}
\label{sec:slot-att}
Following previous studies on describing structured knowledge~\cite{biogen16,sha2017order,table2text17}, we apply a sequence-to-sequence based approach, and incorporate a slot-aware attention to generate the descriptions.

\textbf{Encoder} Given a structured KB input: $L=[(s_1, v_1, (r_1, \hat{r_1})), ..., (s_n, v_n, (r_n, \hat{r_n}))]$, where $s_i$, $v_i$, $r_i$, $\hat{r_i}$ are randomly embedded as vectors $\textbf{s}_i$, $\textbf{v}_i$, $\textbf{r}_i$, $\hat{\textbf{r}}_i$\footnote{We use bold mathematical symbols to denote vector representations for the whole paper.} respectively, we concatenate the vector representations of these fields as $\textbf{l}_i=[\textbf{s}_i, \textbf{v}_i, \textbf{r}_1, \hat{\textbf{r}_1}]$, and obtain $\textbf{L}=[\textbf{l}_1, \textbf{l}_2, ..., \textbf{l}_n]$.

We attempted to apply the average of $\textbf{L}$ as the representation for the input KB. However, such flat representation vectors fail to capture the structured contextual information in the entire KB. Therefore, we apply a bi-directional Gated Recurrent Unit (GRU) encoder~\citep{cho2014learning} on $\textbf{L}$ to produce the encoder hidden states $\textbf{H}=[\textbf{h}_1, \textbf{h}_2, ..., \textbf{h}_n]$, where $\textbf{h}_i$ is a hidden state for $\textbf{l}_i$.

\textbf{Decoder with Slot-aware Attention} The decoder is a forward GRU network with an initial hidden state $\textbf{h}_n$, which is the encoder hidden state of the last token. In order to capture the association between a slot type and its slot value, we design a \textbf{Slot-aware Attention}. 
At each step $t$, we compute the attention distribution over the sequence of input triples. For each triple $i$, we assign it an attention weight: 

\begin{displaymath}
\vspace{-2mm}
\resizebox{.48 \textwidth}{!}{
$e^{t}_{i} = v^{\top}\tanh{\left(W_h \tilde{\textbf{h}}^t + W_s \textbf{s}_i + W_v \textbf{v}_i +  W_c c^{t}_{i} +b_e\right)}$
}
\end{displaymath}
\begin{displaymath}
\alpha ^t= \text{Softmax}\left(e^{t}\right) \ 
\end{displaymath}
where $\tilde{\textbf{h}}^t$ is the decoder hidden state at step $t$. $\textbf{s}_i$ and $\textbf{v}_i$ denote the embedding representations of slot type $s_i$ and slot value $v_i$ respectively. $c^t_{i}=\sum_{k=0}^{t-1}\alpha ^{k}_{i}$ is a coverage vector, which is the sum of attention distributions over all previous decoder time steps and can be used to reduce repetition~\cite{hybridp17}.

The source attention distribution $\alpha^{t}$ can be considered as the contribution of each source triple to the generation of the target word. Next we use $\alpha^t$ to compute two context vectors $\textbf{L}_{s}^{*}$ and $\textbf{L}_{v}^{*}$ as the representation of the slot types and values respectively:
\begin{equation}
\begin{split}
\textbf{L}_{s}^{*} =\sum\nolimits_{i=1}^{n} \alpha^{t}_{i} \textbf{s}_i
\\
\textbf{L}_{v}^{*} =\sum\nolimits_{i=1}^{n} \alpha^{t}_{i} \textbf{v}_i
\end{split}
\label{eq:slotContext}
\end{equation}

At step $t$, the vocabulary distribution $P_{vocab}$ is computed with the context vectors $\textbf{L}_{s}^{*}$, $\textbf{L}_{v}^{*}$ and the decoder hidden state $\tilde{\textbf{h}}^t$, using an affine-Softmax layer:
\begin{displaymath}
P_{vocab} = \text{Softmax}\left(V[\tilde{\textbf{h}}^t;\textbf{L}_{s}^{*};\textbf{L}_{v}^{*}]+b_{vocab}\right)
\label{eq3}
\end{displaymath}

The loss function is computed as:
\begin{displaymath}
\resizebox{.48 \textwidth}{!}{
$Loss = \sum\nolimits_{t} \Big\{-\log{P_{vocab}(y_t)} + \lambda\sum\nolimits_{i}\min{\left(\alpha^t_i,c^t_i\right)} \Big\} $
}
\end{displaymath}
where $P_{vocab}(y^t)$ is the prediction probability of the ground truth token $y_t$. $\lambda$ is a hyperparameter.

\subsection{Table Position Self-attention}
Although the sequence-to-sequence attention model takes into account the information of input triples, it still encodes the structured knowledge as sequential facts while ignoring the correlations between facts. In our task, multiple inter-dependent slots should be described within one sentence. For example, in Table~\ref{tab:example1input}, the sport team \textit{Israel women's national football team} should be described together with \textit{22} matches and \textit{29} goals.
Previous studies~\cite{lin2017structured,vaswani2017attention} applied self-attention on sentence level to capture the correlation between continuous tokens. Inspired by these approaches, we design a new table position based self-attention and incorporate it into the slot-aware attention.


In our task, since most triples are organized in temporal order, we use the row index $r$ and the reverse row index $\hat{r}$ to denote the position information of each triple in the input KB. Given a structured KB as input: $L=[(s_1, v_1, (r_1, \hat{r}_1)), ..., (s_n, v_n, (r_n, \hat{r}_n))]$, we obtain a sequence of row index embeddings $R=[\textbf{r}_{1}^{'}, \textbf{r}_{2}^{'}, ..., \textbf{r}_{n}^{'}]$ with random initialization, where $\textbf{r}_{i}^{'} = [\textbf{r}_i; \hat{\textbf{r}}_i]$. We model the inter-dependencies among slots 
as a latent structure, where for each position $i$ we assume it has a latent in-link and an out-link to denote where it is linked to or from. This assumption is similar to the structure attention applied in~\newcite{18struct}, which assumes each word within a sentence can be a parent node or a child node in a latent tree structure. For each pair of slots $i$ and $j$, we compute the attention score $f_{ij}$ as follows:

\begin{displaymath}
g_{in}=\tanh{\left(W_{in} \textbf{r}_{i}^{'}\right)}
\end{displaymath}
\begin{displaymath}
g_{out}=\tanh{\left(W_{out} \textbf{r}_{j}^{'}\right)}
\end{displaymath}
\begin{displaymath}
f_{ij}= \text{Softmax} \left( g_{in}^{\top}W_g g_{out}\right)
\end{displaymath}
where $W_{in}, W_{out}$, and $W_g$ are learnable parameters.
The attention score will not change during the decoding process. 

$f_{ij}$ can be viewed as the contribution from a context triple $j$ to triple $i$. For each slot $\textbf{s}_i$ and value $\textbf{v}_i$, we obtain a context vector by collecting information from other slot types and their values:

\begin{displaymath}
\textbf{s}_{i}^{*}=\sum\nolimits_{k=1}^{n} f_{ik} \textbf{s}_{k}
\end{displaymath}
\begin{displaymath}
\textbf{v}_{i}^{*}=\sum\nolimits_{k=1}^{n} f_{ik} \textbf{v}_{k}
\end{displaymath} 

We further encode position-aware representation of each slot type and value, and update their context vectors $\textbf{L}_{t}^{*}$ and $\textbf{L}_{v}^{*}$ in Equation~\ref{eq:slotContext} as:

\begin{displaymath}
\textbf{L}_{s}^{*} =\sum\nolimits_{i=1}^{n} \alpha^{t}_{i}\textbf{s}_{i}^{*}
\end{displaymath}
\begin{displaymath}
\textbf{L}_{v}^{*} =\sum\nolimits_{i=1}^{n} \alpha^{t}_{i} \textbf{v}_{i}^{*}
\end{displaymath}

\subsection{Structure Generator}

Traditional sequence-to-sequence models predict a target sequence by only selecting words from a vocabulary with a fixed size. However, in our task, we regard the slot value as a single information unit. Therefore, there is a certain amount of out-of-vocabulary (OOV) words during the test phase. Inspired by the pointer-generator~\cite{copy16,hybridp17}, which is designed to automatically locate particular source words and directly copy them into the target sequence, we design a structure-aware generator as follows.

We first obtain a source attention distribution of all unique input slot values. Since one particular slot value may occur in the structure input for many times, we aggregate the attention weights for each unique slot value $v_j$ from $\alpha_t$ and obtain its aggregated source attention distribution $P_{source}^{j}$ by

\begin{displaymath}
P_{source}^{j} = \sum_{m|v_m=v_j} \alpha^{t}_{m}
\label{eq2}
\end{displaymath}

The \emph{gates} in neural networks act on the signals they receive, and block or pass on information based on its strength. In order to combine two types of attention distribution $P_{source}$ and $P_{vocab}$, we compute a structure-aware gate $p_{gen}\in[0,1]$ as a soft switch between generating a word from the fixed vocabulary and copying a slot value from the structured input:
\begin{multline*}
\resizebox{.47 \textwidth}{!}{
$p_{gen} = \sigma\left( W^{\top}_s \textbf{L}_{s}^{*} + W^{\top}_v \textbf{L}_{v}^{*} + W^{\top}_h \tilde{\textbf{h}}^{t} + W^{\top}_y \textbf{y}^{t-1} + b_{gen}\right)$
}
\end{multline*}
where $\textbf{y}^{t-1}$ is the embedding of the previous generated token at time $t-1$, and $\sigma$ is a Sigmoid function.

The final probability of a token $y$ at time $t$ can be computed by $p_{gen}$, $P_{vocab}$ and $P_{source}$:
\begin{displaymath}
P(y_t) = p_{gen}P_{vocab}+(1-p_{gen})P_{source}
\end{displaymath}
The loss function, combining with the coverage loss~\citep{hybridp17}, is presented as:
\begin{displaymath}
\resizebox{.48 \textwidth}{!}{
$Loss = \sum\nolimits_{t} \Big\{-\log{P(y^{t})}+\lambda\sum\nolimits_{i}\min{\left(\alpha^t_i,c^t_i\right)}\Big\}$
}
\end{displaymath}
where $P(y^t)$ is the prediction probability of the ground truth token $y$. $\lambda$ is a hyperparameter.

\section{Experiments}


\begin{table*}[!htb]
\centering
\small
\setlength\tabcolsep{4pt}
\setlength\extrarowheight{2pt}
\begin{tabularx}{\linewidth}{|>{\hsize=0.7\hsize}X|>{\centering\arraybackslash\hsize=0.8\hsize}X|>{\centering\arraybackslash\hsize=1.4\hsize}X|>{\centering\arraybackslash\hsize=1.3\hsize}X|>{\centering\arraybackslash\hsize=1\hsize}X|>{\centering\arraybackslash\hsize=1\hsize}X|>{\centering\arraybackslash\hsize=0.9\hsize}X|>{\centering\arraybackslash\hsize=0.9\hsize}X|>{\centering\arraybackslash\hsize=1\hsize}X|}
\hline
\textbf{Entity type} & \textbf{\# entity} &\textbf{\# types before filtering}& \textbf{\# types after filtering}  & \textbf{\# slots / sentence} & \textbf{\# words / sentence} & \textbf{\# slots / table} & \textbf{\# words / entity} & \textbf{\# sentence / entity}\\ \hline
Person    & 100,000      & 109 &76     & 1.9   & 16.8    & 8.0     & 70.9 & 4.2                       \\ \hline
Animal    & 6,216        & 30  &12     & 1.3   & 17.1    & 3.2     & 42.2 & 2.5                 \\ \hline
\end{tabularx}
\caption{Data Statistics\label{tabld:sta}}
\end{table*}

\begin{figure*}[!htb]
\centering
\includegraphics[width=\textwidth]{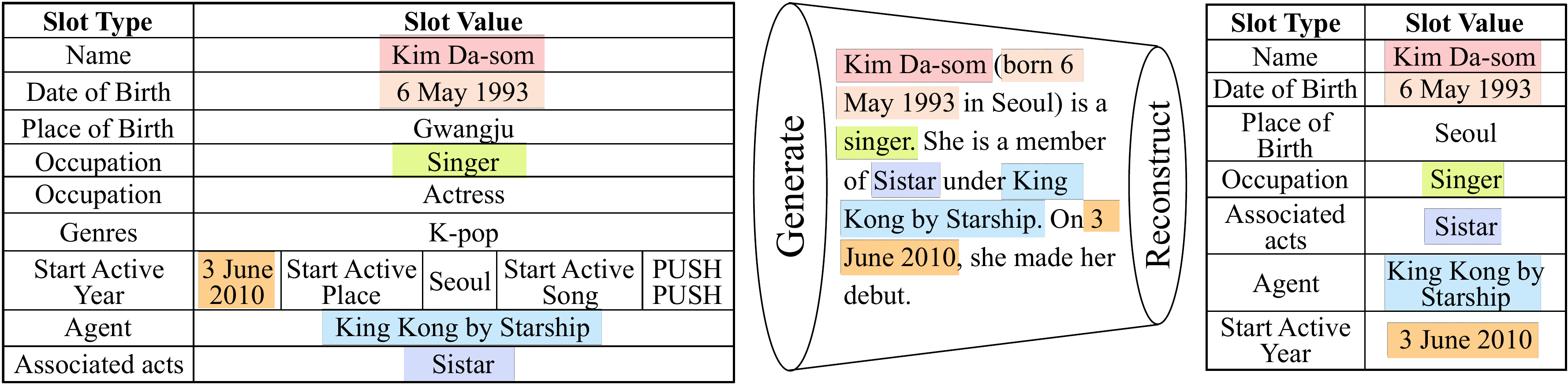}
\vspace{-4mm}
\caption{KB Reconstruction based Evaluation (Scores for the example: Overall Slot Filling P=$\frac{6}{7}$=85.7\%,  R=$\frac{6}{11}$=54.5\%, F1=66.7\%; Inter-dependent Slot Filling P=$\frac{5}{7}$=71.4\%, R=$\frac{5}{9}$=55.6\%, F1=62.5\%) 
}
\vspace{-2mm}
\label{fig:noise}
\end{figure*}
\subsection{Data}


Using person and animal entities as case studies, we create a new dataset based on Wikipedia dump (2018/04/01) and Wikidata (2018/04/12) as follows: (1). Extract Wikipedia pages and Wikidata tables about person and animal entities, and align them according to their unique KB IDs. (2). For each Wikidata table, filter out the slot types of which frequency is less than 3. For each Wikipedia article, use its anchor links (clickable texts in hyperlinks)  to locate all the entities and determine their KB IDs. (3). For each Wikidata table, search each value (including Number, Date) and entity contained in the table in the corresponding Wikipedia article according to its KB ID, and remove the values and entities which cannot be found in the corresponding Wikipedia article.
(4). For each Wikipedia article, remove the sentences which contain no values, and remove sentences which only contain entities that do not exist in the Wikidata table. 
The remaining sentences will be taken as ground-truth reference descriptions. 
(5). Index the row numbers for each slot type according to their orders in the Wikidata table. The ground-truth structured KB is then created. (6). Build a fixed vocabulary for the whole corpus of ground-truth descriptions and label the words with frequency $<5$ as OOV.








We further randomly shuffle and split the dataset into training (80\%), development (10\%) and test (10\%) subsets for person and animal entities respectively. Table~\ref{tabld:sta} shows the detailed statistics. Compared with the Wikibio dataset used in previous studies~\cite{biogen16,sha2017order,table2text17}, which contains one sentence only as the ground-truth description, our dataset contains multiple sentences to cover as many facts as possible in the input structured  KB. It makes 
the generation task more challenging, practical and interesting.

\subsection{Evaluation Metrics}

We apply the standard BLEU~\citep{Bleu02}, METEOR~\cite{denkowski2014meteor}, and ROUGE~\citep{lin2004rouge} metrics to evaluate the generation performance, because they can measure the content overlap between system output and ground-truth and also check whether the system output is written in sufficiently good English.

In addition, we 
can also consider natural language as the most expressive way for knowledge transmission via a \emph{noisy channel}. 
If we are able to reconstruct the input KB from the generated description, our generator achieves a 100\% success rate at knowledge propagation. We propose a \emph{KB reconstruction based metric} as follows: for each entity, construct a KB from the generated paragraph, and compute precision, recall and F-score by comparing it with the input KB from two aspects: 
(1). \textbf{Overall Slot Filling}: If a pair of slot type and its slot value exists in both of the reconstructed KB and the input KB, it's considered as a correct slot. 
(2). \textbf{Inter-dependent Slot Filling}: If a row that consists one or multiple slot types and their slot values exist in both of the reconstructed KB and the input KB, it's considered as a correct row.


If the same slot/row is correctly described multiple times in the system generation output, it's only counted as correct once, i.e., redundant descriptions will be penalized. 
This metric is further illustrated in Figure~\ref{fig:noise}. It's similar to the relation extraction based generation evaluation metric proposed by~\cite{data2docu17} and entity/event extraction based metric proposed by~\cite{Whitehead18,Lu18imagecaption}. They compared automatic Information Extraction results from the reference description and the system generation output. However, the performance of state-of-the-art open-domain slot filling~\cite{wu2010open,fader2011identifying,min2012ensemble,xu2013open,angeli2015leveraging, bhutani2016nested,Yu2017} is still far from satisfactory to serve as an automatic extraction tool for evaluating generation results. Therefore for the pilot study in this paper we manually reconstruct KBs from the generation output for evaluation. Notably none of the above automatic metrics is sufficient to capture adequacy, grammaticality and fluency of the generated descriptions. However extrinsic metrics such as system purpose and user task are expensive, while cheaper metrics such as human rating do not correlate with extrinsic metrics~\cite{Gkatzia2015}. Moreover the task we address in this paper requires essential domain knowledge for a human user to assess the generated descriptions.

\subsection{Baseline Models}


We compare our approach with the following models: (1). \textbf{Seq2seq attention model}~\cite{atten15}. We concatenate slot types and values as a sequence, e.g., $\{$\textit{Name, Silvi Jan, Sports team, ASA Tel Aviv University, Hapoel Tel Aviv F.C.} ...$\}$ for Table~\ref{tab:example1input}, and apply the sequence to sequence with attention model to generate a description. (2). \textbf{Pointer-generator}~\cite{hybridp17} which introduces a soft switch to choose between generating a word from the fixed vocabulary and copying a word from the input sequence. Here, we concatenate all slot values as the input sequence, e.g., $\{$\textit{Silvi Jan, ASA Tel Aviv University, Hapoel Tel Aviv F.C.} ...$\}$ for Table~\ref{tab:example1input}. (3). \textbf{Pointer-generator + slot type attention} which incorporates the slot type attention (Section~\ref{sec:slot-att}) into the pointer-generator. We use the sequence of (slot type, slot value) pairs as input, e.g., $\{$\textit{(Name, Silvi Jan), (Sports team, ASA Tel Aviv University), (Sports team, Hapoel Tel Aviv F.C.)} ...$\}$ for Table~\ref{tab:example1input}.


\subsection{Hyperparameters}

Table~\ref{tab:Hyperparameter} shows the hyperparameters of our model.

\begin{table}[htb!]
\setlength\tabcolsep{4pt}
\setlength\extrarowheight{3pt}
\small
\centering
\begin{tabularx}{\linewidth}{|>{\hsize=1.1\hsize}X|>{\centering\arraybackslash\hsize=0.9\hsize}X|}
\hline
\textbf{Parameter}                       & \textbf{Value} \\ \hline
Vocabulary size (|s|+|v|)                & 46,776          \\ \hline
Value\textbackslash{type} embedding size & 256            \\ \hline
Position embedding  size                 & 5              \\ \hline
Slot embedding size                     & 522            \\ \hline
Decoder hidden  size                     & 256            \\ \hline
Coverage loss $\lambda$          & 1.5              \\ \hline
Optimization                             & Adam~\citep{kingma2014adam}           \\ \hline
Learning rate                            & 0.001          \\ \hline
\end{tabularx}
\caption{Hyperparameters\label{tab:Hyperparameter}}
\end{table}
\subsection{Results and Analysis}
\label{sec_analysis}

Table~\ref{table:methodcomparison} shows the performance of various models with standard metrics. We can see that our attention mechanisms achieve consistent improvement. We conduct paired t-test between our proposed model and all the other baselines on 10 randomly sampled subsets.  The differences are statistically significant with $p \leq 0.016$ for all settings. 

As shown in Table~\ref{table:kbconstruction} and Table~\ref{table:kbconstructiondep}, the KBs reconstructed from models with these two attention mechanisms achieve much higher quality. 

\begin{table*}[!htb]
\centering
\small
\setlength\tabcolsep{4pt}
\setlength\extrarowheight{2pt}
\begin{tabularx}{\linewidth}{|>{\hsize=1.4\hsize}X|>{\centering\arraybackslash\hsize=0.9\hsize}X|>{\centering\arraybackslash\hsize=1\hsize}X|>{\centering\arraybackslash\hsize=0.9\hsize}X|>{\centering\arraybackslash\hsize=0.9\hsize}X|>{\centering\arraybackslash\hsize=1\hsize}X|>{\centering\arraybackslash\hsize=0.9\hsize}X|}
\hline
\multirow{2}{*}{\textbf{Model}} & \multicolumn{3}{c|}{\textbf{Person}} & \multicolumn{3}{c|}{\textbf{Animal}} \\ \cline{2-7} 
 & \textbf{BLEU} & \textbf{METEOR} & \textbf{ROUGE}  & \textbf{BLEU} & \textbf{METEOR} & \textbf{ROUGE}  \\ \hline
Seq2seq & 11.3 & 16.9 & 28.8  & 5.8 & 11.5 & 20.5  \\ \hline
Pointer & 17.2 & 21.1 & 37.4  & 6.6 & 13.7 & 37.8 \\ \hline
+Type & 23.1 & 22.2 & 39.5 & \textbf{17.2} & \textbf{17.3} & 42.8  \\ \hline
+Type \& Position & \textbf{23.2} & \textbf{23.4}& \textbf{42.0} & 14.8 & 17.2 & \textbf{45.0}\\ \hline
\end{tabularx}
\caption{Generation Performance based on Standard Metrics \%)\label{table:methodcomparison}}
\vspace{-2mm}
\end{table*}

\begin{table}[!htb]
\small
\centering
\setlength\tabcolsep{4pt}
\setlength\extrarowheight{2pt}
\begin{tabularx}{\linewidth}{|>{\hsize=2.8\hsize}X|>{\centering\arraybackslash\hsize=0.7\hsize}X|>{\centering\arraybackslash\hsize=0.7\hsize}X|>{\centering\arraybackslash\hsize=0.7\hsize}X|>{\centering\arraybackslash\hsize=0.7\hsize}X|>{\centering\arraybackslash\hsize=0.7\hsize}X|>{\centering\arraybackslash\hsize=0.7\hsize}X|}
\hline
\multirow{2}{*}{\textbf{Model}} & \multicolumn{3}{c|}{\textbf{Person}} & \multicolumn{3}{c|}{\textbf{Animal}} \\ \cline{2-7} 
 & \textbf{P} & \textbf{R} & \textbf{F1}  & \textbf{P} & \textbf{R} & \textbf{F1} \\ \hline
Seq2seq & 74.6 & 29.3 & 42.0 & 82.5 & 27.8 & 41.6 \\ \hline
Pointer & 72.6 & 56.4 & 62.8 & 58.5 & 37.5 & 45.7\\ \hline
+Type & 75.9 & 58.8 & 66.3  & 65.9 & 63.8 & 64.8 \\ \hline
+Type \& Position & \textbf{76.3} & \textbf{62.7} &\textbf{68.8}  & \textbf{73.4} & \textbf{71.8} & \textbf{72.6} \\ \hline
\end{tabularx}
\caption{Overall Slot Filling Precision (P), Recall (R), F-score (F1) (\%)\label{table:kbconstruction}}
\vspace{-2mm}
\end{table}

\begin{table}[!htb]
\small
\centering
\setlength\tabcolsep{4pt}
\setlength\extrarowheight{2pt}
\begin{tabularx}{\linewidth}{|>{\hsize=2.8\hsize}X|>{\centering\arraybackslash\hsize=0.7\hsize}X|>{\centering\arraybackslash\hsize=0.7\hsize}X|>{\centering\arraybackslash\hsize=0.7\hsize}X|>{\centering\arraybackslash\hsize=0.7\hsize}X|>{\centering\arraybackslash\hsize=0.7\hsize}X|>{\centering\arraybackslash\hsize=0.7\hsize}X|}
\hline
\multirow{2}{*}{\textbf{Model}} & \multicolumn{3}{c|}{\textbf{Person}} & \multicolumn{3}{c|}{\textbf{Animal}} \\ \cline{2-7} 
 & \textbf{P} & \textbf{R} & \textbf{F1}  & \textbf{P} & \textbf{R} & \textbf{F1} \\ \hline
Seq2seq & 74.7 & 30.0 & 43.4 & 82.5 & 27.9 & 41.7 \\ \hline
Pointer & 73.0 & 56.4 & 63.6 & 57.7 & 37.2 & 45.2\\ \hline
+Type & 75.8 & 58.9 & 66.3  & 66.3 & 64.2 & 65.2 \\ \hline
+Type \& Position & \textbf{77.2} & \textbf{63.5} & \textbf{69.7} & \textbf{72.6} & \textbf{71.0} & \textbf{71.8}  \\ \hline
\end{tabularx}
\caption{Inter-dependent Slot Filling Precision (P), Recall (R), F-score (F1) (\%)\label{table:kbconstructiondep}}
\vspace{-2mm}
\end{table}

\begin{figure}[!htb]
\centering
\includegraphics[width=0.5\textwidth]{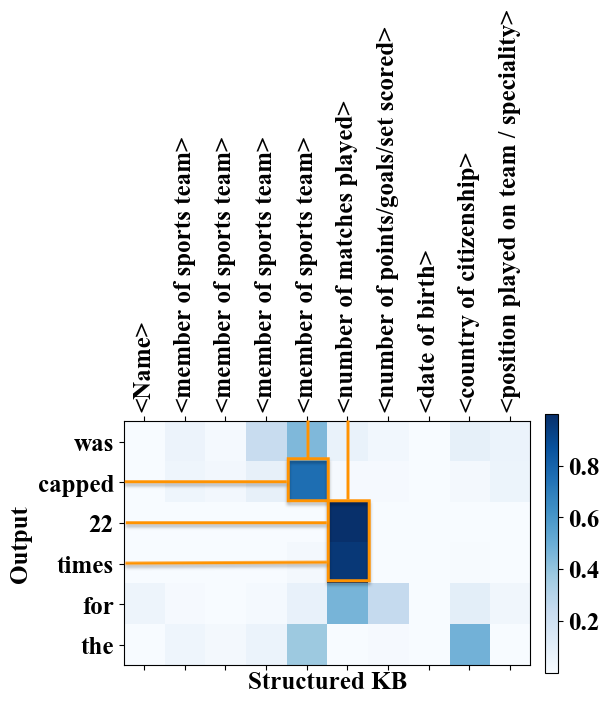}
\vspace{-10mm}
\caption{Slot Type Attention Visualization (Context words strongly associated with certain slot types receive high weights, e.g., \emph{capped} to describe \emph{member of sports team}, and \emph{times} to describe \emph{the number of matches played}. )
}
\label{fig:type}
\end{figure}

\begin{figure*}[!htb]
\centering
\includegraphics[width=\textwidth]{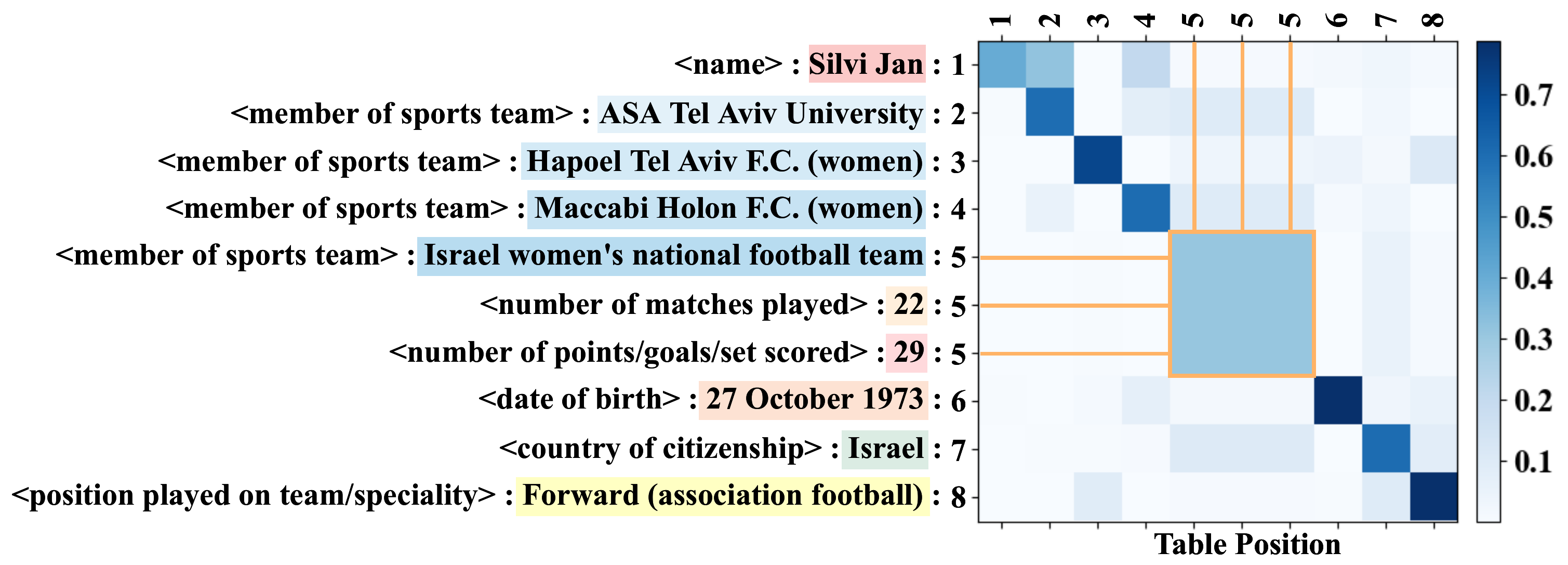}
\vspace{-10mm}
\caption{Table Position Self Attention Visualization (the highlighted inter-dependent slots appear in the same row and the same sentences, and thus they receive the same high weight.)}
\label{fig:self}
\end{figure*}

Figure~\ref{fig:type} and Figure~\ref{fig:self} visualize the attentions applied to the walk-through example in Table~\ref{tab:example1input}.

\paragraph{Impact of Slot-aware Attention.} The same string can be filled into various slots of multiple types. For example, dates, ages, the number of matches and goals can all be presented as numbers. The pointer network often mistakenly mixes them up. For example, it produces ``\emph{24 September 1979 was born 3 October 1903 in \textbf{17} on \textbf{33} October 1906}'', where \textit{33} should be the number of matches and \textit{17} should be the number of goals. In contrast our model with slot type attention correctly generates ``\emph{he made \textbf{33} appearances and scored \textbf{17} goals}''. In addition, as mentioned earlier, the pointer network often produces redundant slot values because it loses control of slot types, e.g., ``\emph{He was born in the city of \textbf{Association football}. In the late 1990s he was appointed manager of the \textbf{Association football} team of the team.}''. 

\paragraph{Impact of Table Position Attention.} The table position attention successfully captures inter-dependent slots, such as a membership with a sports team and its corresponding number of matches and games: ``\emph{Bill Sampy ... who played for \textbf{Sheffield United F.C.} \textbf{41} times.}''; ``\emph{Giancarlo Antognoni ... he was also a member of the \textbf{Italy national football team} at the \textbf{1982 FIFA World Cup}.}''.

\paragraph{Remaining Challenges.} 
Some remaining errors are trivial to fix, such as fixing a country name to its adjective form when it appears right before a position slot (e.g.,  \textit{Italian professional Association football player} instead of  \textit{Italy professional Association football player}). The KB reconstruction recall of person entities is relatively low mainly because we don't have enough training data for some rare slot types. 

Contextual words generated by the LM introduces some incorrect facts, especially temporal expressions. For example, the generator does not have the commonsense knowledge that football players could not play before they were born: ``\emph{Aleksei Gasilin ( born \textbf{1 March 1996} ) is a Russian Association football Forward (association football). He made his professional debut in the Russian Second Division in \textbf{1992} for Russia national under-19 football team. }''. Similarly, a football player would probably not be still active when he was already 72 years old: ``\emph{Basil Rigg ( born \textbf{12 August 1926} ) is a former Australian rules football Rigg played for the Perth Football Club in the Western Australia cricket team from \textbf{1998} to \textbf{1998}.}''.


Our approach sometimes fails to detect person gender so as to generate incorrect pronouns. 
For animal entities, human writers are able to elaborate more details. For example, human writes the specific endemic places for \emph{Brown treecreeper}: ``\emph{The bird endemic to eastern Australia has a broad distribution occupying areas from \textbf{Cape York Queensland} throughout \textbf{New South Wales} and \textbf{Victoria} to \textbf{Port Augusta} and the \textbf{Flinders Ranges} \textbf{South Australia}.}'' while our system is only able to cover the generic location information ``\emph{It is endemic to \textbf{Australia}.}'' from the input KB.

\section{Related work}




Our task is similar to the WebNLG challenge generating text from DBPedia data~\cite{gardent2017creating}. Previous approaches on generating natural language sentences from structured input KB can be divided into two categories: the first is to induce templates and then fill appropriate content into slots~\cite{kukich1983design,cawsey1997natural,prob110,link13,konstas2013global,flanigan2016generation}. These methods can generate high-quality descriptions but heavily rely on information redundancy to create templates. The second category is to directly generate a sequence of words using language model~\cite{belz2008automatic, chen2008learning,liang2009learning,prob110,P12-1039,concept12,konstas2013global, concept13,sta16} or deep neural networks~\cite{sutskever2011generating,wensclstm15, checklist16, Alignment16, P17-1017, data2docu17,P18-2042, song2018graph}. 
Several studies~\cite{biogen16,17onebio,kaffee2018learning,kaffee2018mind,table2text17,sha2017order} generate a person's biography from an input structure, which are closely related to our task. However, instead of modeling the input structure as a sequence of facts and generating one sentence only, we introduce a table position self-attention, inspired from structure attention~\cite{lin2017structured,kim2017structured,vaswani2017attention,shen2018disan, shen2018bidirectional}, to capture the dependencies among facts and generate a paragraph to describe all facts. 



In contrast to some recent work on converting structured Abstract Meaning Representation~\cite{Banarescu2013} into natural language~\cite{Pourdamghani2016,Flanigan2016}, our task requires us to capture inter-dependent relation links in a knowledge base and use them to generate multiple sentences in most cases. Our work is also related to attention mechanisms for sequence-to-sequence generation~\cite{atten15,Alignment16,ma2017improving}. Different from previous studies, our task requires the slot type and slot value to appear in the generated sentences in pairs. Thus we design a slot-aware attention to obtain two context vectors for both slot type and slot value simultaneously. 
To deal with OOV 
words, we use a structure generator, which is similar to the pointer-generator networks~\cite{pointer15,rare15, gulcehre2016pointing, hybridp17} and copy mechanism~\cite{copy16}.

\section{Conclusions and Future Work}

We develop an effective generator to produce a natural language description about an input knowledge base. Our experiments show that two attention mechanisms focusing on slot type and table position advance state-of-the-art on this task, and provide a KB reconstruction F-score up to 73\%. We propose a new KB reconstruction based evaluation metric which can be used for other knowledge-driven NLG tasks such as news image/video captioning. In the future, we aim to address the remaining challenges as summarized in Section~\ref{sec_analysis}, and tackle the setting where multiple facts of the same slot type are not presented in temporal order in the input KB. 
We also plan to extend the framework to cross-lingual cross-media generation, namely to produce a foreign language description or an image/video about the KB.






\section*{Acknowledgments}
This work was supported by the U.S. DARPA AIDA Program No. FA8750-18-2-0014 and U.S. ARL NS-CTA No. W911NF-09-2-0053. The views and conclusions contained in this document are those of the authors and should not be interpreted as representing the official policies, either expressed or implied, of the U.S. Government. The U.S. Government is authorized to reproduce and distribute reprints for Government purposes notwithstanding any copyright notation here on.

\bibliography{acl2018}
\bibliographystyle{acl_natbib}

\end{document}